\documentclass[runningheads]{llncs}
\usepackage[T1]{fontenc}
\usepackage{amsmath}
\usepackage{newtxtext}
\usepackage[varvw]{newtxmath}
\usepackage{graphicx}
\usepackage{booktabs}
\usepackage{color}
\usepackage{url}
\usepackage{hyperref}

\begin{document}

\title{ChildAI-Gov: An Evidence-Linked AI System for Assessing EdTech Policies Against Children's Digital Rights}
\titlerunning{ChildAI-Gov: Evidence-Linked AI Governance for EdTech}
\authorrunning{N. Sigalapalli et al.}

\author{Naveen Sigalapalli\inst{1} \and Huizhi Liang\inst{1} \and Clara Crivellaro\inst{1}}

\institute{School of Computing, Newcastle University, Newcastle upon Tyne, United Kingdom \\
\email{n.sigalapalli2@newcastle.ac.uk}, \email{huizhi.liang@newcastle.ac.uk}, \email{clara.crivellaro@newcastle.ac.uk} \\
Corresponding author: Naveen Sigalapalli (\email{n.sigalapalli2@newcastle.ac.uk})}

\maketitle
{\renewcommand{\thefootnote}{}\footnotetext{Accepted at the 10th Euro-China Conference on Intelligent Data Analysis and Applications (ECC 2026). The final authenticated version will appear in Springer's Smart Innovation, Systems and Technologies (SIST) series.}}

\begin{abstract}
We present ChildAI-Gov, an explainable AI system for assessing whether education-technology (EdTech) policies respect children's digital rights. Generative AI is reaching children's education faster than its governance can keep pace, yet checking a tool against children's rights remains manual, expert-dependent, and slow. We propose the Child Rights Impact Score (CRIS), an evidence-linked measure across five dimensions: safety, privacy, transparency, autonomy, and non-discrimination. ChildAI-Gov reads an EdTech tool's privacy or AI policy and scores each dimension by combining a large-language-model judgement (gpt-4o-mini) with a deterministic rule check. A retrieval-augmented generation (RAG) component grounds every score in the governing provisions, retrieved from a curated knowledge base of children's rights frameworks (the EU AI Act, UNICEF guidance, and others), so each score is tied to the evidence behind it and an auditor can trace which provision it rests on. We also construct AIPolicy-EdTech, a dataset of 57 real EdTech policies. On it, the five dimensions cohered as a single measure (Cronbach's $\alpha$ = 0.79), and grounding made scoring stricter while preserving rankings; an ungrounded baseline was far more lenient, passing nearly half the policies where ChildAI-Gov passed none. It offers evidence-linked decision support on documented commitments, not observation of real-world behaviour or determinations of legal compliance.

\end{abstract}

\section{Introduction}
Generative AI has moved into children's daily lives with unusual speed. A recent survey of 25,592 children aged nine to sixteen across twenty European countries found that roughly seven in ten use generative AI, with schoolwork the most common use~\cite{staksrud2026}. It is therefore not a peripheral novelty in children's education but an everyday presence that collects their data, mediates their learning, and in some cases converses with them directly.

This ordinary presence carries governance consequences. The education-technology (EdTech) tools through which children encounter generative AI process personal data, inform automated decisions, and increasingly hold open-ended conversations with young users, engaging children's rights to privacy, to protection from commercial exploitation, to information, and to safety. A growing body of regulation and guidance speaks to these concerns: the UK Information Commissioner's Office (ICO) Age Appropriate Design Code~\cite{ico2020}, the Online Safety Act 2023~\cite{osa2023}, the EU AI Act, which identifies children as requiring heightened protection~\cite{euaiact}, and UNICEF's Policy Guidance on AI for Children~\cite{unicef2021}. Each addresses overlapping but non-identical concerns in dense legal language.

Whether a tool meets these expectations is hard to establish. Doing so requires reading its privacy or AI policy, cross-referencing several frameworks at once, and exercising legal judgement about what the text commits to: accurate when performed by experts, but slow, costly, and unscalable. A recent expert audit of generative AI in education illustrates both value and cost, producing detailed rights-based findings through product walkthroughs, policy analysis, and consultation, but demanding substantial specialist effort per tool~\cite{atabey2025}. Because expert audit of this kind cannot scale to the number of tools in use, a widening gap opens between the tools warranting scrutiny and the capacity to provide it, and most tools reaching children are therefore never assessed. A second gap concerns the children themselves: these policies are written for adults and lawyers, yet exercising a right depends on understanding it, which motivates the safeguarded child-facing component we introduce below.

\begin{figure}[t]
\centering
\includegraphics[width=\textwidth]{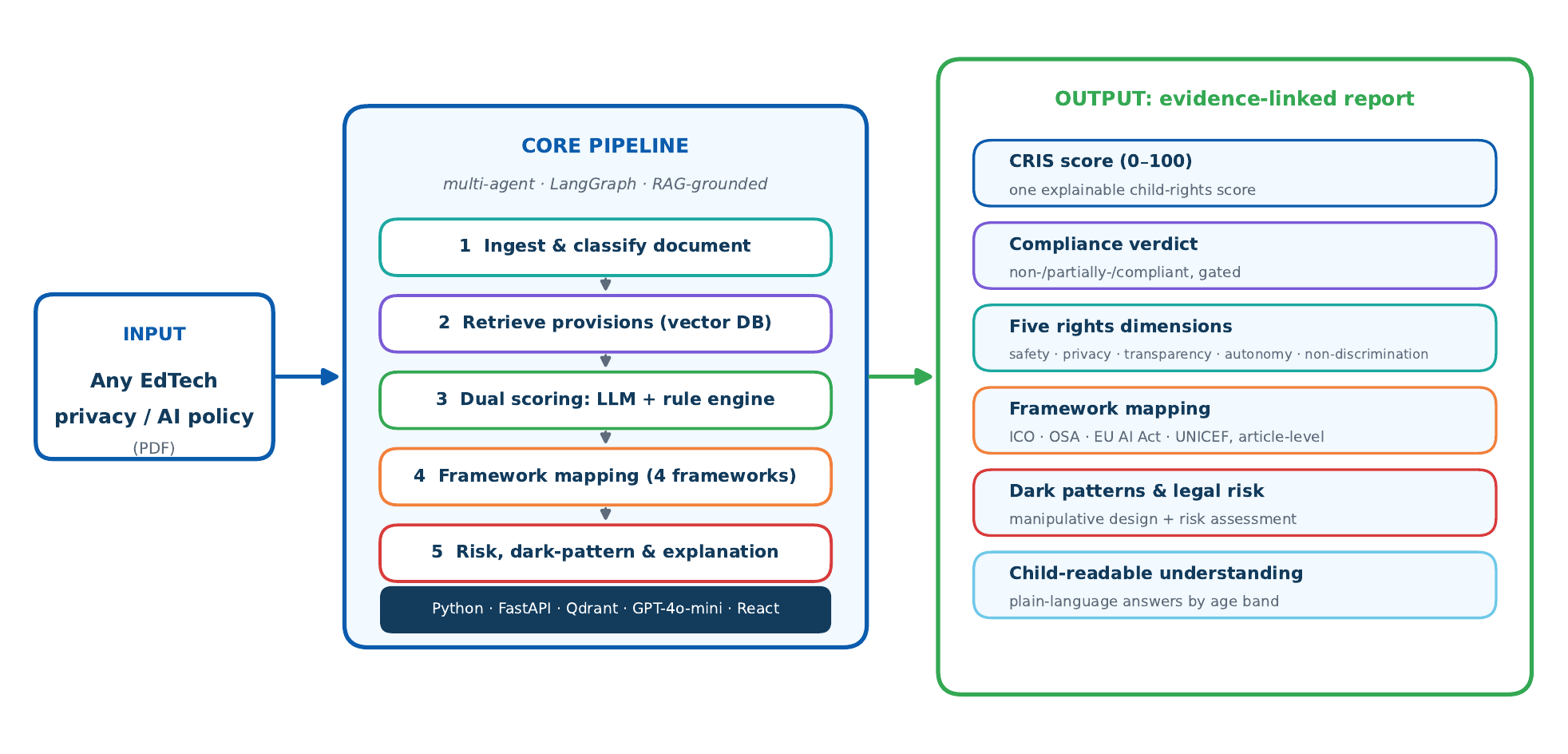}
\Description{Flow diagram of the ChildAI-Gov pipeline: an EdTech policy is ingested and classified, framework provisions are retrieved from a knowledge base, five rights dimensions are scored by a language model and a rule engine, the policy is mapped against four frameworks, and a risk, dark-pattern and explanation stage produces an evidence-linked report.}
\caption{Overview of the ChildAI-Gov pipeline. Any EdTech policy is ingested and classified; provisions are retrieved from a framework knowledge base; five dimensions are scored by a language model and a rule engine; the policy is mapped against four frameworks; and risk, dark patterns, and explanations are produced, yielding an evidence-linked report.}
\label{fig:pipeline}
\end{figure}

This paper presents ChildAI-Gov, which addresses this gap by reading a policy and producing an evidence-linked, child-rights assessment. Fig.~\ref{fig:pipeline} gives an overview: any EdTech policy enters the pipeline and a structured report emerges. Our aim is to determine whether automated, evidence-linked analysis of policy text can provide scalable child-rights assessment of EdTech tools against four child-rights and governance frameworks spanning the UK, the EU, and international guidance, while establishing honestly the boundary of what document-based assessment can and cannot achieve. Concretely, we ask whether framework provisions can be mapped to EdTech policy text so that every assessment is traceable to supporting evidence, how language-model judgement can be combined with deterministic detection of binding-obligation language to keep that evidential basis inspectable, and what proportion of expert-identified concerns is detectable from policy text alone.

Our contribution is not a new retrieval or language-model technique but an evidence-linked governance system, together with the artefacts needed to apply and test it. Specifically, we propose: (i) the Child Rights Impact Score (CRIS), an evidence-linked measure in which every dimension score is tied, through a retrieval-augmented generation (RAG) component, to named framework provisions rather than to model recall, so a human auditor can see exactly which provision each judgement rests on and where a policy satisfies or misses it; (ii) a hybrid dual-scoring method that combines semantic language-model judgement with a deterministic rule engine so their disagreement is inspectable; (iii) AIPolicy-EdTech, a dataset of 57 real EdTech privacy and AI policies on which we calibrate the system, showing quantitatively that unaided language-model scoring is systematically lenient and that grounding corrects it while preserving rankings; and (iv) a measurable expert boundary: taking an independent expert audit as a reference, we quantify how much of what experts flag can be detected from the policy text alone, and how much depends instead on observing the product in use. To our knowledge this combination, applied to multi-framework child-rights governance of generative AI in EdTech and paired with a safeguarded child-facing assistant, has not been demonstrated before.

\section{Related Work}
This work draws on three areas developed largely in parallel: children's rights in digital environments, the frameworks translating those rights into obligations, and the automated analysis of policy text.

\subsection{Children's Digital Rights and Child-Centred AI}
Children's rights in digital environments are grounded in the UN Convention on the Rights of the Child and, more recently, General Comment No. 25, which clarifies how existing rights apply in digital settings~\cite{gc25}. UNICEF's Policy Guidance on AI for Children translates them into requirements spanning safety, privacy, non-discrimination, transparency, and development~\cite{unicef2021}. A recurring theme, directly relevant to the dimensions we assess, is that child-focused regulation conceptualises children primarily as objects of protection rather than active rights-holders, so participation and agency are less fully operationalised than safety~\cite{5rights2023}. Most directly relevant is a recent child-rights audit of generative AI in EdTech~\cite{atabey2025}. Using mixed sociolegal methods, including product walkthroughs, policy analysis, and consultation, it assessed widely used tools against the UNCRC and identified concerns spanning privacy, commercial exploitation, safety, and information. That audit provides a rare expert reference standard, and its methodology exposes a dependency the automation literature tends to leave implicit: many of its most serious findings, such as a chatbot giving unsafe responses or trackers activating during use, were established by observing the product rather than reading its policy. We treat that dependency as measurable rather than assumed.

\subsection{Automated Analysis of Privacy and Policy Text}
The automated analysis of privacy policies is a mature field. The OPP-115 corpus of 115 annotated policies provided a foundation for subsequent systems~\cite{wilson2016}. Polisis segments a policy and classifies each segment against a taxonomy of privacy practices using neural classifiers~\cite{harkous2018}. It links predictions to the supporting segment, offering evidence-based explainability; it composes rule-based logic over classifier outputs, an early example of combining learned and deterministic components; and it answers free-form questions from a policy. Its authors are explicit that it is not a substitute for the legal document but a means of providing initial evidence for experts. We share this decision-support orientation and attention to traceable evidence, but differ in domain and audience: Polisis targets general privacy practices for adults, whereas we target child-rights provisions across multiple frameworks with an interface designed for children.

A limitation of segment classification is that it labels what a policy discusses without modelling how its statements relate. PoliGraph responds by extracting a knowledge graph linking data types, purposes, and recipients, which exposes internal contradictions that segment-level labels cannot~\cite{poligraph}. A separate strand examines children's privacy specifically, largely under the United States Children's Online Privacy Protection Act (COPPA). Zimmeck and colleagues combined machine-learning analysis of policies with static analysis of application code, finding only around a third of thousands of policies contained the statements COPPA requires~\cite{zimmeck2017}. Reyes and colleagues took a behavioural approach, running thousands of children's applications in an instrumented environment and finding a majority potentially in violation through third-party tracking~\cite{reyes2018}. Together these establish a point we build upon: what a policy states and what a product does can diverge, and detecting each requires different methods. Closest in spirit to our scoring approach is a computable gap assessment of governance guidance~\cite{meng2025}, though it analyses one framework to find where guidance is underspecified, whereas we analyse real EdTech policies against several frameworks, and it includes no child-facing component.

\subsection{Positioning}
The individual techniques we use are established in this work; our contribution is their integration and application to multi-framework child-rights governance of generative AI in EdTech, with a safeguarded child-facing assistant and evaluation against an independent expert audit. We position these prior systems as conceptual and architectural baselines rather than quantitative ones: a direct experimental comparison is outside our scope because their objectives, target frameworks, and datasets differ from ours (general adult privacy practice or single-framework COPPA compliance, versus multi-framework child-rights assessment). We do not report direct quantitative comparisons with Polisis or the gap-assessment work because their task definitions, output taxonomies, datasets, and evaluation protocols differ substantially from ours. Our quantitative evaluation therefore compares ChildAI-Gov (the grounded pipeline) against a language-model-only baseline on the same dataset and task, isolating the contribution of grounding and dual scoring.

\section{Method}
\subsection{Overview and Design Rationale}
\begin{figure}[t]
\centering
\includegraphics[width=\textwidth]{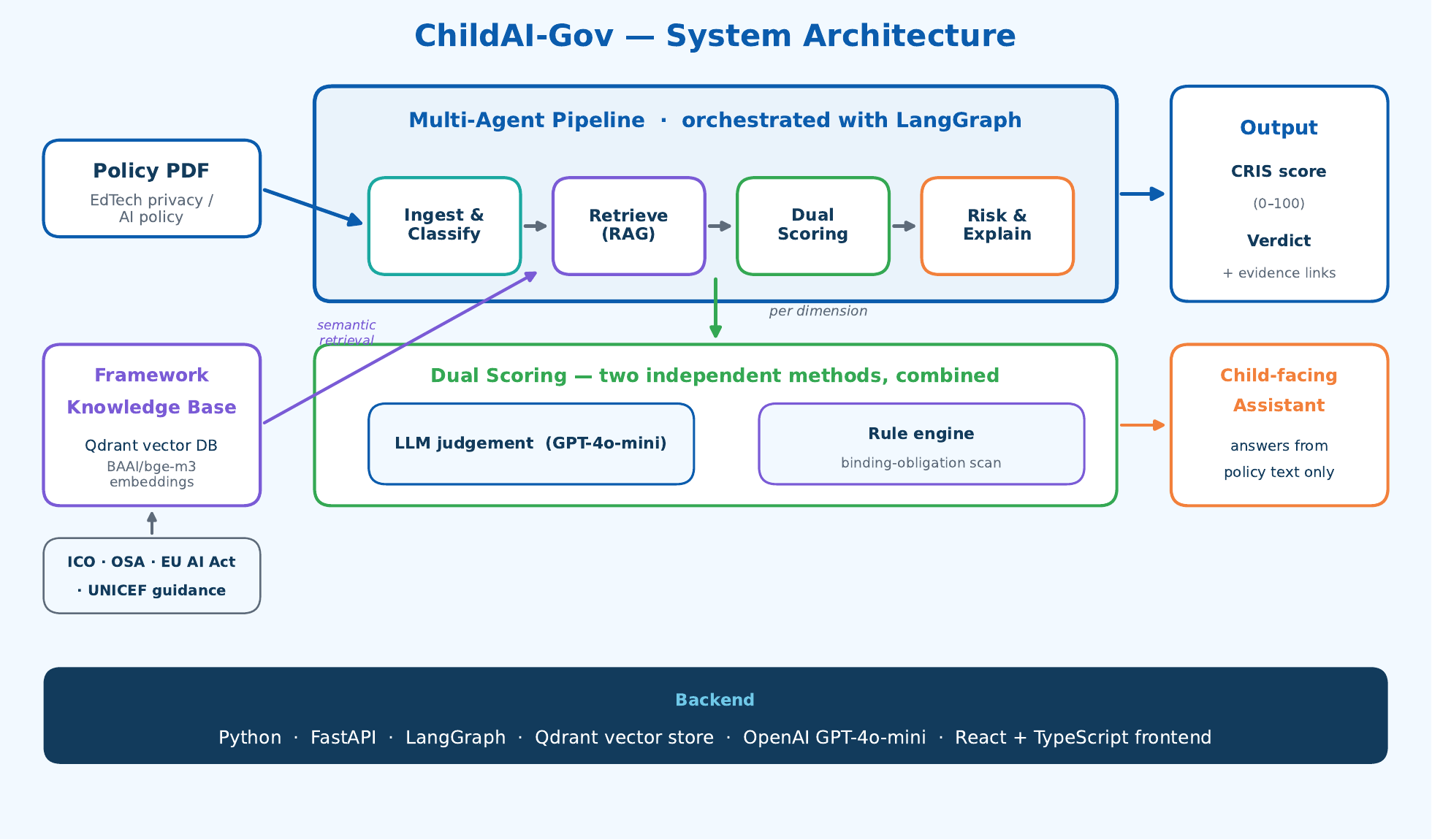}
\Description{System architecture diagram showing a modular, agent-orchestrated LangGraph pipeline with retrieval-augmented grounding from a separated framework knowledge base and a dual-scoring stage.}
\caption{System architecture: a modular, agent-orchestrated pipeline implemented with LangGraph, with retrieval-augmented grounding from a separated framework knowledge base and dual scoring.}
\label{fig:architecture}
\end{figure}

ChildAI-Gov analyses a single EdTech privacy or AI policy and produces a structured, evidence-linked assessment of how far its documented commitments align with child-rights expectations. It is organised as a modular, agent-orchestrated pipeline implemented using LangGraph (Fig.~\ref{fig:architecture}), with distinct responsibilities: a classification stage, a retrieval stage, a scoring stage, a risk-assessment stage, and an explanation and report-generation stage. Each stage is independently inspectable, a precondition for evidence-linked assessment that a monolithic prompt could not meet. The pipeline uses Python with FastAPI, a Qdrant vector store, and gpt-4o-mini. We state the scope precisely: the system performs policy-level rights screening on documented commitments, not legal compliance determination, and reasons over document evidence, not observed real-world behaviour.

\subsection{Document Ingestion and Classification}
A policy is supplied as a PDF, from which text is extracted and stored. The system classifies it as a privacy policy, AI policy, guidance document, or binding governance policy by matching characteristic terms in the opening text. The type carries a penalty weight adjusting framework scoring: a non-binding guidance document should not be judged by the same enforceability standard as a mandatory policy. Classification defaults conservatively to privacy policy, since these form the dataset majority and a conservative default limits the impact of misclassification.

\subsection{Knowledge Base and Retrieval}
The system draws on a curated knowledge base of child-rights provisions extracted from the four frameworks. Each provision is stored with its framework, article or standard identifier, description, and the dimension it primarily serves, and is embedded using the BAAI/bge-m3 model~\cite{bgem3} into a dedicated Qdrant collection held separately from any policy text. The separation is deliberate: holding provisions and policy chunks in one collection risks the framework knowledge base being overwritten as policies are ingested, and a separate collection makes that failure structurally impossible. When scoring a dimension, the system queries this collection and retrieves the most relevant provisions by semantic similarity above a threshold of 0.50. Retrieval is semantic rather than lexical, so a clause can be related to a requirement despite different wording, which matters because policies rarely adopt regulatory vocabulary verbatim. The purpose of this retrieval-augmented generation (RAG) step is grounding: rather than relying on what the language model happens to recall about regulation, each dimension is judged against the actual provisions retrieved for it, and those provisions are carried forward into the report. This is the mechanism behind the evidence-linking that defines the system, and it is what the dark-pattern and risk gates, the dimension scores, and the child-facing explanations are all built on.

\subsection{Governance Dimensions}
The platform assesses five dimensions: safety, privacy, transparency, autonomy, and non-discrimination. These were arrived at by mapping the encoded provisions to the child-rights concern each expresses, and grouping recurring concerns into interpretable categories (Table~\ref{tab:dimensions}). Safety, privacy, and transparency are each supported by provisions from three or four frameworks, and are therefore grounded in a convergence of instruments. Autonomy and non-discrimination are supported more sparingly, a genuine feature of the regulatory landscape rather than a defect of the mapping; their limited framework support is reflected in their lower weighting.

\begin{table}[t]
\centering
\caption{Derivation of the five governance dimensions from framework provisions.}
\label{tab:dimensions}
\begin{tabular}{ll}
\toprule
\textbf{Dimension} & \textbf{Contributing provisions} \\
\midrule
Safety & OSA Part 3, s47; UNICEF Req 3; EU AI Act Art 5, 9; ICO Std 13 \\
Privacy & ICO Std 2, 7, 8, 10, 12; UNICEF Req 5 \\
Transparency & ICO Std 4; EU AI Act Art 13, 50; UNICEF Req 6 \\
Autonomy & UNICEF Req 7; ICO Std 1, 3 \\
Non-discrimination & EU AI Act Art 10; UNICEF Req 2 \\
\bottomrule
\end{tabular}
\end{table}

\subsection{Dual Scoring}
The central methodological choice is that each dimension is scored twice, by language-model semantic judgement and by a deterministic rule-based check, and the two combined. They respond to different features of the text: a language model can judge whether a commitment is substantive but is opaque and variable, while keyword matching is transparent and reproducible but cannot distinguish a genuine commitment from the mere presence of vocabulary. Retaining both scores makes their disagreement visible and informative.

The language-model judgement is a prompted, zero-shot evaluation using gpt-4o-mini at temperature zero. For each dimension the model receives the dimension's criteria, the retrieved provisions, and the policy text, and grades commitments as absent (1--3), mentioned but vague (4--5), defined with mechanisms (6--7), or fully enforced with accountability (8--10). It is instructed to evaluate strictly and not default to the middle of the scale. The rule-based check matches curated dimension-specific keywords against the policy text, scaled so a policy with none of the expected terms scores low and one with all scores high. The two are combined as a weighted sum, the language model contributing 0.6 and the rule check 0.4, favouring semantic judgement while retaining a deterministic anchor that limits the influence of model variability. Both component scores are stored alongside the hybrid, so disagreement remains visible. Each dimension score is also retained together with the provisions retrieved for it and the policy passages they were matched against, so the report states, for every dimension, which provision is satisfied or missing and on what textual basis. This is what we mean by an evidence-linked assessment: a human auditor can follow each score back to the specific regulation text and the policy wording behind it, rather than being asked to trust an opaque number.

\subsection{Composite Score and Verdict}
The five hybrid dimension scores are combined into a single Child Rights Impact Score on a 0 to 100 scale:
\begin{equation}
\mathrm{CRIS} = 10 \times \sum_{i=1}^{5} w_i H_i
\end{equation}
where $H_i$ is the hybrid score for dimension $i$ on a 0 to 10 scale and $w_i$ its weight, the weights summing to one: safety and privacy at 0.25 each, transparency at 0.20, autonomy and non-discrimination at 0.15 each. These weights are a transparent design choice rather than an optimised parameter, giving greatest weight to the dimensions most densely represented across frameworks and most central to child protection. A verdict combines the score with two gates: a detected manipulative-design (dark) pattern yields non-compliant regardless of score, since no documented compliance offsets manipulation directed at children; a low score with high assessed regulatory risk also yields non-compliant. Because the CRIS derives partly from a language model, its numeric value carries run-to-run variation; the verdict, less sensitive to small numeric changes, is treated as the primary output.

\subsection{Child-Facing Assistant}
To extend transparency to the children whose data is at stake, the platform includes an assistant answering a child's questions about a policy. It answers only from the policy text, reducing fabrication risk, and is governed by a safeguarding instruction set: it offers no legal advice; it does not falsely reassure a child their data is safe when the policy indicates otherwise; and on questions about a child's rights or data it directs them to a trusted adult. This last rule reflects expert consultation during design: answering a child's rights question without reference to the adults who can act on it risks isolating the child from the support the right depends on.

\subsection{Dataset}
The platform was applied to \emph{AIPolicy-EdTech}, a dataset of 57 EdTech and related privacy policies that we constructed, selected purposively rather than as a statistical sample. The dataset spans child-directed and early-years services, mainstream classroom tools, and UK and EU services chosen to match the jurisdictions assessed. It includes widely used tools such as MagicSchool AI and Grammarly, the two tools examined in the expert audit, so output could be compared against an independent reference. Policies were collected in the versions publicly available in early 2026 and are English-language; length ranges from about 3,200 to 96,000 characters (mean 29,000, median 24,800). Twelve exceed the 40,000-character reading window, a documented limitation. This supports assessment across policy types and quality levels; it is not a representative survey of the market.

\section{Results and Analysis}
\subsection{Assessment on the AIPolicy-EdTech Dataset}
No policy was assessed as fully compliant; 33 were non-compliant and 24 partially compliant, with CRIS from 31.0 to 60.9 (mean 43.7, median 42.8, SD 6.8). Scores span a genuine range rather than clustering: 17 policies below 40, 30 between 40 and 49, 9 between 50 and 59, one at 60 or above. That none approaches the compliant threshold is itself a finding.

\subsection{A Worked Example}
To make the output concrete, Table~\ref{tab:worked} shows the dimension-level assessment of a representative mid-range policy (CRIS 42.3, verdict partially compliant). The two scoring components agree closely on safety but diverge sharply on transparency: the policy uses the expected vocabulary, so the rule component scores it 6.0, while the language model, judging whether the disclosures are genuinely intelligible to a child, scores it 1.0. The hybrid score of 3.0 records that tension rather than concealing it, and the accompanying report cites the ICO and EU AI Act transparency provisions against which the shortfall was identified. No dark pattern was detected, so the score rather than a gate determines the verdict. The example illustrates the system's intended use: not a single number, but a per-dimension, provision-linked account a non-specialist can read and act on.

\begin{table}[t]
\centering
\caption{Dimension-level assessment of a representative policy (CRIS = 42.3; verdict: partially compliant). The LLM and rule components are on a 0--10 scale; the hybrid is their weighted combination (0.6 and 0.4). The weighted hybrid scores sum to 4.23, giving a CRIS of 42.3.}
\label{tab:worked}
\begin{tabular}{lcccc}
\toprule
\textbf{Dimension} & \textbf{LLM} & \textbf{Rule} & \textbf{Hybrid} & \textbf{Weight} \\
\midrule
Safety & 5.0 & 5.0 & 5.0 & 0.25 \\
Privacy & 3.0 & 4.25 & 3.5 & 0.25 \\
Transparency & 1.0 & 6.0 & 3.0 & 0.20 \\
Autonomy & 4.0 & 5.25 & 4.5 & 0.15 \\
Non-discrimination & 5.0 & 6.25 & 5.5 & 0.15 \\
\bottomrule
\end{tabular}
\end{table}

\subsection{Agreement Between Scoring Components}
Across the 285 dimension assessments (57 policies $\times$ 5 dimensions), the two components fall within one point on 52\% of assessments and within two on 81\%, and are weakly correlated overall (Pearson's $r = 0.25$), indicating they are far from redundant. The divergence is clearest on transparency, where keyword presence rates policies highly (mean rule score 6.1) while the language model, judging whether disclosures are genuinely child-accessible, rates the same policies low (mean 3.3), the clearest evidence that the dual design captures something a single method would miss.

\subsection{Reliability, Internal Structure, and Sensitivity}
The composite shows acceptable internal consistency (Cronbach's $\alpha = 0.79$), with a correlated protection core of safety, privacy, and non-discrimination and two more loosely coupled dimensions. An ablation recomputing every policy under language-model-only, rule-only, and hybrid scoring gave mean CRIS of 41.0, 47.7, and 43.7, but altered the verdict for only one of 57 policies, since the dark-pattern and risk gates dominate. Equal dimension weights changed the mean by about one point with no verdict changes, so results are not an artefact of the chosen weights. We are explicit that internal consistency establishes only that the dimensions cohere; the CRIS is an engineered screening metric, not an externally validated measurement instrument.

\subsection{Effect of Provision Grounding}
To isolate the effect of grounding we re-scored all 57 policies with retrieval disabled, so the model judged each dimension against its criteria alone. Grounding made assessment stricter: mean CRIS fell from 45.4 to 43.7, with 43 of 57 policies scoring lower, 10 higher, and 4 unchanged. This is expected, since judging against named provisions is more demanding than against a general description. Grounding nonetheless preserved structure: the score sets correlate strongly ($r = 0.93$), no verdict changed, and internal consistency was essentially identical ($\alpha = 0.79$ versus 0.78). Grounding shifts the level of the scores and ties them to identifiable provisions without altering conclusions; we adopt it as the default.

\subsection{Comparison Against a Language-Model-Only Baseline}
\begin{figure}[t]
\centering
\includegraphics[width=0.75\textwidth]{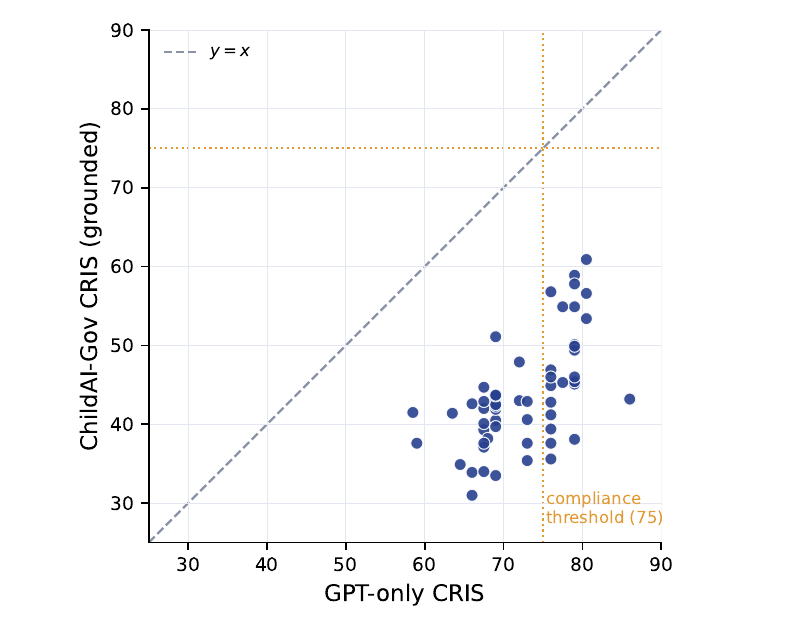}
\Description{Scatter plot of GPT-only CRIS against ChildAI-Gov grounded CRIS for each policy; all points lie below the line y equals x, and 25 points exceed the compliance threshold of 75 on the GPT-only axis while none do on the ChildAI-Gov axis.}
\caption{GPT-only versus ChildAI-Gov (grounded) CRIS for each policy in the AIPolicy-EdTech dataset. Every policy lies below the $y=x$ line; 25 cross the compliance threshold (75) under naive scoring, none under ChildAI-Gov.}
\label{fig:baseline}
\end{figure}

To assess what grounding adds over naive prompting we implemented a GPT-only baseline: the same model, temperature, and window, prompted in a single call to score all five dimensions directly, without grounding, rule component, or structured grading. It scored every policy higher than ChildAI-Gov, by a mean of 28.9 points (baseline mean 72.5, range 58.5--86.0; ChildAI-Gov 43.7). Applying the compliance threshold to the raw baseline scores, 25 of 57 policies would reach the compliant range, whereas ChildAI-Gov placed none there (Fig.~\ref{fig:baseline}). The two nonetheless rank policies similarly (Spearman's $\rho = 0.65$), so the model has genuine signal about relative quality. What grounding adds is not a different ordering but a different calibration, correcting the leniency of unaided prompting. For a screening tool the direction matters: certifying weak policies is the more damaging error, and the one ChildAI-Gov avoids.

\subsection{Child-Facing Assistant}
We evaluated the assistant on 36 questions across six policies against manually verified ground truth. It referred children to a trusted adult in all 36 scenarios and produced no false reassurances under the evaluation protocol. Factual accuracy rose from 44\% to 69\% once retrieval coverage of long policies was corrected (exact McNemar, $p = 0.004$). We characterise this as a promising preliminary safety evaluation rather than definitive validation: the sample is modest ($n = 36$) and used a single language-model judge.

\subsection{Comparison Against an Expert Audit}
We conducted a limited directional comparison against the independent expert child-rights audit of two tools~\cite{atabey2025}. Roughly two-thirds of the expert findings rest on behavioural or consultative evidence, actual data flows, product behaviour, or stakeholder consultation, that no document-based tool can observe. This locates where automated assessment assists expert review and where it cannot replace it. We are careful not to describe this as validation against experts: it is a directional comparison over two tools, not an inter-rater study over the corpus.

\subsection{Ongoing Validation}
\label{subsec:ongoing}
Two further evaluations are in progress and are reported here as planned protocols rather than completed results. For retrieval quality, we are assembling a sample of 100 retrieved provision-policy pairs, each independently judged for relevance, from which we will report precision@$k$ and mean average precision. For external validation, we are designing a small blinded study in which several child-rights experts each score a subset of policies on the same five dimensions, blind to system output; inter-rater agreement and correlation with the CRIS will provide the external validation the composite score currently lacks and permit principled recalibration of the verdict thresholds.

\section{Discussion}
The dual scoring's value is chiefly the granular, evidence-linked dimension score, which shows where a policy is strong and weak while only occasionally driving the verdict. Grounding makes scores stricter and traceable while preserving rankings and every verdict. Against the language-model-only baseline the value becomes concrete: naive prompting is lenient enough to certify nearly half the dataset, yet ranks policies similarly, so the engineering contributes calibration, not ranking.

Two findings appear to carry beyond this system. The first is that unaided language-model scoring is not merely noisy but systematically lenient, erring in the direction that matters most: it certifies weak policies rather than flagging good ones. As automated governance tooling proliferates, that asymmetry deserves attention in its own right. The second is that the boundary of document-based assessment is measurable rather than merely assertable, which converts a caveat into a result and gives a concrete target for the hybrid approaches needed to close the gap.

Limitations. The platform performs policy-level rights assessment of documented commitments, not enacted behaviour, and does not determine legal compliance. The external check is limited to a directional comparison on two tools, which is no substitute for an expert panel scoring policies with inter-rater agreement; that study is in progress (Section~\ref{subsec:ongoing}). We do not compare against other published automated systems beyond a language-model baseline, nor evaluate retrieval quality with ranking metrics in this version. The dataset is 57 English-language policies, twelve exceeding the reading window. The rule engine is keyword-based, a deliberate transparency trade-off, and the 0.6/0.4 weighting and 0.50 retrieval threshold are manually selected, tested through sensitivity analysis but not learned.

\section{Conclusion and Future Work}
We presented ChildAI-Gov, a system that assesses an EdTech policy against child-rights provisions drawn from four child-rights and governance frameworks, produces an evidence-linked Child Rights Impact Score and verdict, and extends its transparency to children through a safeguarded assistant. The curated knowledge base links each dimension assessment to retrieved framework provisions and supporting policy evidence; the dual-scoring method combines two genuinely non-redundant components; the composite score is internally consistent and markedly better calibrated than naive prompting, which would have certified nearly half the dataset where ChildAI-Gov certified none; and comparison against an independent expert audit establishes that roughly two-thirds of expert findings rest on evidence no document-based tool can observe.

Beyond the retrieval and expert-panel studies described in Section~\ref{subsec:ongoing}, which will provide external validation of the CRIS and recalibration of its thresholds, further directions include a direct quantitative comparison against published systems such as Polisis once a shared task and dataset are established, replacing the keyword rule engine with richer extraction, extending the dataset beyond English, and, most substantively, pairing document screening with lightweight behavioural probes so automated triage and observed behaviour combine rather than compete.

\begin{credits}
\subsubsection{\ackname} The authors thank the School of Computing at Newcastle University, and Dr Ay\c{c}a Atabey for guidance on the trusted-adult safeguarding design.

\subsubsection{\discintname} The authors have no competing interests to declare that are relevant to the content of this article.
\end{credits}


\begin{thebibliography}{99}
\bibitem{staksrud2026} Staksrud, E., et al.: European Children's Use and Understanding of Generative AI: EU Kids Online 2026. London School of Economics (2026)
\bibitem{ico2020} Information Commissioner's Office: Age Appropriate Design: A Code of Practice for Online Services (2020)
\bibitem{osa2023} UK Parliament: Online Safety Act 2023 (2023)
\bibitem{euaiact} European Parliament and Council: Regulation (EU) 2024/1689 (Artificial Intelligence Act) (2024)
\bibitem{unicef2021} UNICEF: Policy Guidance on AI for Children, Version 2.0 (2021)
\bibitem{gc25} UN Committee on the Rights of the Child: General Comment No. 25 on Children's Rights in Relation to the Digital Environment (2021)
\bibitem{5rights2023} 5Rights Foundation: Child Rights by Design (2023)
\bibitem{atabey2025} Atabey, A., Sylwander, K., Livingstone, S.: A Child Rights Audit of Generative AI in EdTech. Digital Futures for Children / 5Rights Foundation (2025)
\bibitem{wilson2016} Wilson, S., Schaub, F., Dara, A.A., Liu, F., Cherivirala, S., Leon, P.G., Andersen, M.S., Zimmeck, S., Sathyendra, K.M., Russell, N.C., Norton, T.B., Hovy, E., Reidenberg, J., Sadeh, N.: The Creation and Analysis of a Website Privacy Policy Corpus. In: Proceedings of the 54th Annual Meeting of the Association for Computational Linguistics (Volume 1: Long Papers), pp. 1330--1340. Association for Computational Linguistics, Berlin (2016). \doi{10.18653/v1/P16-1126}
\bibitem{harkous2018} Harkous, H., Fawaz, K., Lebret, R., Schaub, F., Shin, K.G., Aberer, K.: Polisis: Automated Analysis and Presentation of Privacy Policies Using Deep Learning. In: 27th USENIX Security Symposium. USENIX Association (2018)
\bibitem{poligraph} Cui, H., Trimananda, R., Markopoulou, A., Jordan, S.: PoliGraph: Automated Privacy Policy Analysis Using Knowledge Graphs. In: 32nd USENIX Security Symposium. USENIX Association (2023)
\bibitem{zimmeck2017} Zimmeck, S., Wang, Z., Zou, L., Iyengar, R., Liu, B., Schaub, F., Wilson, S., Sadeh, N., Bellovin, S.M., Reidenberg, J.: Automated Analysis of Privacy Requirements for Mobile Apps. In: Network and Distributed System Security Symposium (NDSS 2017). Internet Society (2017)
\bibitem{reyes2018} Reyes, I., Wijesekera, P., Reardon, J., Elazari Bar On, A., Razaghpanah, A., Vallina-Rodriguez, N., Egelman, S.: ``Won't Somebody Think of the Children?'' Examining COPPA Compliance at Scale. Proc. Priv. Enhancing Technol. \textbf{2018}(3), 63--83 (2018). \doi{10.1515/popets-2018-0021}
\bibitem{meng2025} Meng, W.: Computable Gap Assessment of Artificial Intelligence Governance in Children's Centres. arXiv:2601.04216 (2025)
\bibitem{bgem3} Chen, J., Xiao, S., Zhang, P., Luo, K., Lian, D., Liu, Z.: BGE M3-Embedding: Multi-lingual, Multi-functionality, Multi-granularity Text Embeddings Through Self-Knowledge Distillation. arXiv:2402.03216 (2024)
\end{thebibliography}
\end{document}